# Machine Learning-Based Research on the Adaptability of Adolescents to Online Education

Mingwei,Wang[1], Sitong Liu[2]

[1]College of Humanities&Social Sciences,HZAU,Wuhan,China
[2]GSE, University of Pennsylvania, Philadelphia, USA

[1]14512340123@qq.com

**Abstract.** With the rapid advancement of internet technology, the adaptability of adolescents to online learning has emerged as a focal point of interest within the educational sphere. However, the academic community's efforts to develop predictive models for adolescent online learning adaptability require further refinement and expansion. Utilizing data from the "Chinese Adolescent Online Education Survey" spanning the years 2014 to 2016, this study implements five machine learning algorithms－logistic regression, K-nearest neighbors, random forest, XGBoost, and CatBoost－to analyze the factors influencing adolescent online learning adaptability and to determine the model best suited for prediction. The research reveals that the duration of courses, the financial status of the family, and age are the primary factors affecting students' adaptability in online learning environments.Additionally, age significantly impacts students' adaptive capacities. Among the predictive models, the random forest, XGBoost, and CatBoost algorithms demonstrate superior forecasting capabilities, with the random forest model being particularly adept at capturing the characteristics of students' adaptability.

## 1. Introduction

In the digital era, the evolution of information technology has catalyzed significant innovation in educational paradigms, with online learning emerging as a pivotal segment within the field of education[1]. The 50th Statistical Report on Internet Development in China, published in 2022, indicates a substantial growth in the user base of online education in China, expanding from 110 million in 2015 to 377 million by 2022. Despite the evident advantages of online education, the industry confronts numerous challenges[2].

Adaptability to online learning is defined as the learner's capacity to actively modify their approach in response to changes in personal and environmental conditions during the process of knowledge acquisition via the internet, thereby aligning their development with the learning context and achieving educational objectives[3]. It is one of the critical factors influencing the development of online education. Although current academic research extensively covers various aspects of adolescent online learning adaptability, including individual, familial, scholastic, and community dimensions, there is a notable absence of in-depth analysis on how these factors interrelate and collectively affect adolescents' acceptance of online education. Moreover, prevailing research often relies on descriptive statistics and correlation analysis to interpret the connections between these factors and online learning adaptability, with a scarcity of studies constructing predictive models based on these elements. The limitations of this methodological approach have hindered our ability to deeply comprehend and anticipate the trends in adolescent adaptability to online education. Consequently, leveraging certain factors to predict the level of adaptability of adolescents to online education is of profound significance.

Accordingly, this study aims to explore and assess the factors influencing the adaptability of adolescent students to online learning and to construct corresponding predictive models, thereby providing a scientific basis and data support for educational practice and policy formulation.

This research refines and organizes the broad concept of online learning adaptability and the dimensions that may affect adolescent online learning adaptability. By establishing a dataset that encompasses 13 features and one target column, an in-depth analysis of the personal background, educational environment, and online learning situation of adolescent students has been conducted. Methodologically, this paper employs univariate and multivariate analysis techniques to provide a comprehensive description and analysis of the dataset. Univariate analysis probes the distribution of individual features, while subsequent multivariate analysis unveils the complex interrelationships between features, particularly highlighting the impact of gender differences and academic majors on the level of adaptability to online learning. In terms of model establishment and evaluation, this paper utilizes five machine learning algorithms: logistic regression, K-nearest neighbors, random forest, XGBoost, and CatBoost.

Ultimately, this study identifies the model most suited for predicting student adaptability to online learning and ensures the robustness and generalizability of the model through further optimization and adjustment. The research outcomes not only enrich the theoretical understanding of online learning adaptability but also offer practical guidance for educators and policymakers. This contributes to the advancement of personalized and precise educational services, enhancing educational quality and promoting educational equity.

## 2. Related work

Adolescent students, as the primary beneficiaries of online education, are inevitably influenced in their learning outcomes by the transformation of their learning environments and the evolution of their learning strategies. This shift encompasses not only the relocation of physical spaces but also profound changes in cognitive processes and the ways in which knowledge is acquired. Consequently, a comprehensive assessment of learning outcomes must account for the integrated effects of these dynamic changes. In this context, the importance of learning adaptability becomes increasingly pronounced. For adolescents, what factors influence their acceptance of online education, and how can these factors be leveraged to predict their receptivity to online learning?

In general, the factors affecting adaptability to online education can be broadly categorized into four dimensions: individual, familial, scholastic, and community.

Individual factors, such as a student's ethnicity[4], are closely related to online learning adaptability; personal attributes, particularly psychological health status[5] and self-efficacy[6], significantly influence students' adaptability to online learning. Regarding the scholastic environment, the school setting plays a pivotal role in adolescents' adaptation to online education. The quality of teaching and course information[7] indirectly affects learning adaptability, exerting a notable positive influence. At the familial level, the accumulation of parental social, cultural, and economic[8] capital fosters students' adaptability to online learning. Additionally, the family provides a robust social support system, thereby enhancing students' adaptability and academic resilience in online learning[9]. The nature of the community in which students reside also impacts their adaptability to online learning, primarily manifesting as differences between urban and rural communities[10].

Although current academic research on the adaptability of adolescents to online learning has extensively covered multiple dimensions, including the individual, family, school, and community, there remains a need for in-depth analysis on how these factors interact and collectively influence adolescents' acceptance of online education. Moreover, most studies lack systematic theoretical guidance, making it difficult for research findings to accumulate and provide coherent guidance for educational practice. Some studies tend to adopt a single-disciplinary perspective, which limits the possibility of a comprehensive understanding of online learning adaptability. Furthermore, most research tends to use descriptive statistics and correlation analysis to elucidate the connections between various factors and online learning adaptability, with few studies constructing predictive models based on these factors. The limitations of this research method restrict researchers' in-depth insights and predictive capabilities regarding the developmental trends of adolescents' adaptability to online education.

Therefore, based on existing research, this paper will use the social ecological systems theory as a framework, integrating perspectives from education, data science, sociology, and other disciplines. Through univariate analysis, it systematically investigates the distribution of various characteristics. Further multivariate analysis reveals the complex relationships between characteristics and identifies the model most suitable for predicting the adaptability of adolescents to online learning.

## 3. Data Collection and Feature Analysis

*3.1 Feature Description*
Features are the attributes or characteristics that describe data samples, offering input information for models to make predictions or classifications. In machine learning, features often serve as independent variables, capturing differences and patterns among samples. The selection and extraction of features are crucial for model performance; optimal features enhance accuracy and generalizability. This study employs features such as gender, age, educational level, and internet type to delineate students' backgrounds, educational settings, and online learning circumstances for adaptability prediction.

The target variable, also known as the dependent variable, is the entity that machine learning models aim to predict or classify. In supervised learning tasks, the target variable contains the true values or categories that the model seeks to learn and predict. In this study, the level of adaptability serves as the target variable, an indicator of students' adaptability within the online educational context. The model's task is to predict or classify students' adaptability levels based on given features, offering guidance and recommendations for educators and policymakers.

**3.2 Univariate Analysis Results**
In the univariate analysis, the gender feature reveals that males and females constitute 55% and 45% of the total population, respectively. Regarding the type of educational institution, non-governmental institutions account for a significant proportion of 68.3%. The age distribution of respondents, mainly between 11 and 25 years old, reflects a general trend in the age distribution of the surveyed population. The class duration predominantly ranges from 1 to 3 hours, indicating a preference and acceptance level for course duration among respondents. The majority of respondents are engaged in higher education levels at schools and universities. In terms of financial condition, most respondents are in a moderate financial state, potentially influencing their degree and manner of participation in online education. The use of mobile data and phones is notably high at 57.7% and 84.1%, respectively, underscoring the widespread application of mobile networks and devices in online education. The adaptability level shows that approximately 51.9% of respondents indicate a moderate level of adaptability, 39.8% a low level, and the remainder a high level, providing crucial insights into the adaptability of the surveyed individuals. Collectively, these univariate analysis results offer vital clues and perspectives for an in-depth understanding of respondents' characteristics and behaviors.

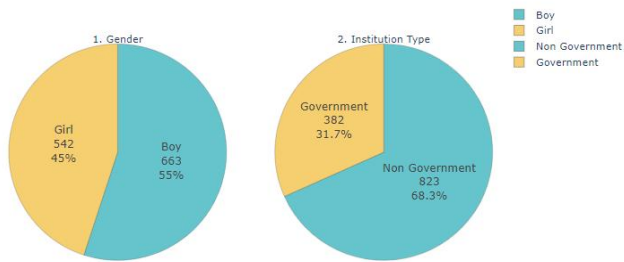
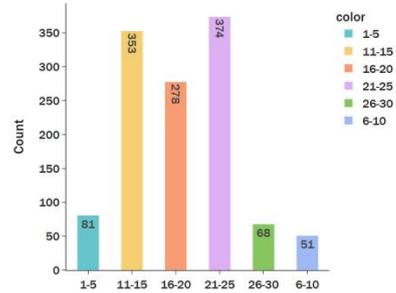

Gender & Institution Type                    Age

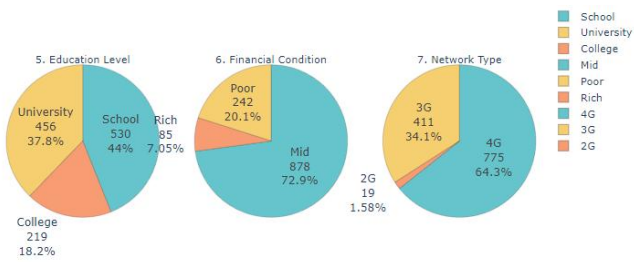
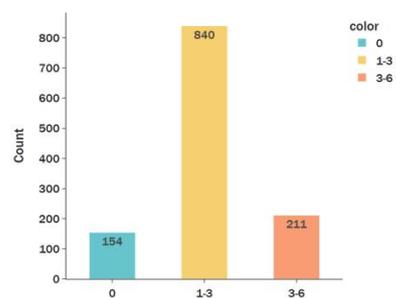

Education Level & Financial Condition & Network Type

Class Duration

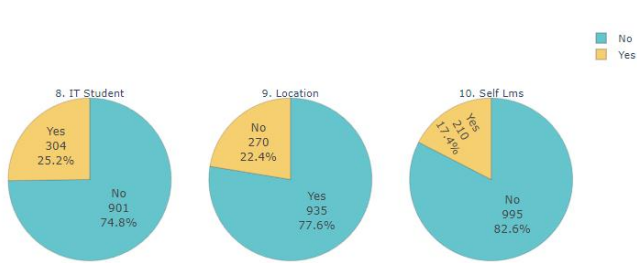
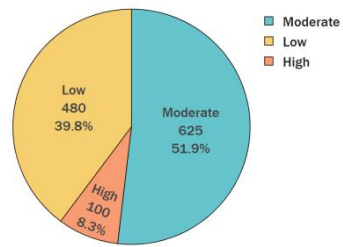

IT Student & Location & Self Lms

Adaptivity Level

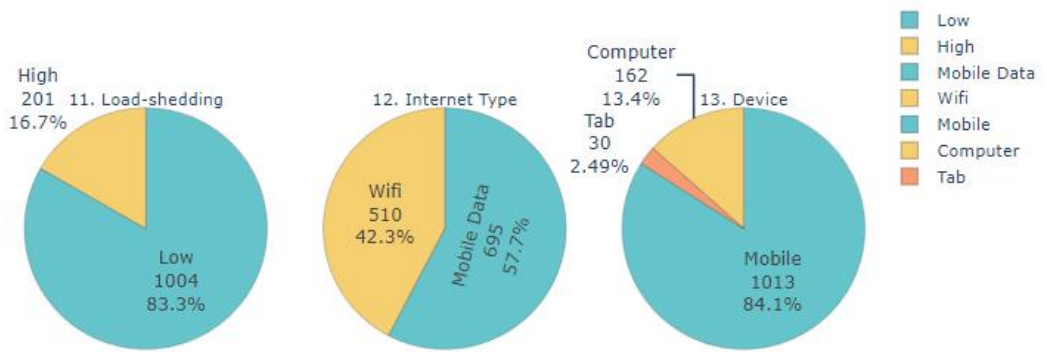

Load-shedding & Internet Type & Device

**Figure 1.** Visualization of Univariate Analysis Results from the Dataset

**4. Model Establishment and Evaluation**

*4.1 Data Preprocessing*
During the data preprocessing phase, features often require encoding to convert categorical or textual data into a numerical form that models can process. In this case, label encoding was utilized to transform the categorical values of each feature into numerical codes. This ensures that the model can correctly comprehend the relationships between features and process them appropriately.

To train and evaluate the machine learning models, the dataset is typically divided into training and testing sets. The training set is used to train the model, while the testing set is employed to assess model performance. Here, the dataset was split into training and testing sets, with 70% of the data allocated for training, encompassing 843 samples, and 30% for testing, comprising 362 samples. The partitioning process ensures that the model has sufficient data for effective learning and evaluation during both training and testing.

*4.2 Model Selection*
In this study, to construct and evaluate the online education adaptability prediction model, we have carefully selected five distinct machine learning algorithms. These include logistic regression, K-nearest neighbors, random forest, XGBoost, and CatBoost.

Logistic regression is a classic classification algorithm suitable for binary classification problems. It is based on a linear regression model and employs a logistic function to convert continuous predictor variables into probability values. Logistic regression offers advantages in simplicity and interpretability and performs well with large-scale data.

The K-nearest neighbors algorithm is an instance-based learning method that classifies based on the labels of the K closest neighbors in the feature space. This algorithm does not require prior model training but instead makes predictions directly from the training data, offering good flexibility.

Random forest is an ensemble learning algorithm that enhances prediction accuracy and robustness by constructing multiple decision trees and averaging their results. Random forest can handle a large number of input features and performs well with missing data and outliers.

XGBoost is a gradient-boosted tree algorithm that iteratively trains decision tree models, using the residuals of previous models as the target for new models, thereby gradually improving predictive performance. XGBoost excels in processing structured data and high-dimensional features and has a fast training speed.

CatBoost is a gradient-boosted tree-based algorithm specifically designed for classification problems and capable of automatically handling categorical features and missing values. CatBoost demonstrates good robustness when dealing with large-scale data and highly sparse features.

By selecting these five different machine learning algorithms, we can comprehensively compare their performance in predicting online education adaptability and determine the most suitable model for solving the problem.

*4.3 Cross-Validation Results and Comparison of Models*
In this study, the five selected machine learning models underwent 5-fold cross-validation, and their average accuracy on the validation set was calculated. Figure 2 illustrates the accuracy scores of the classifiers. The results indicate that the random forest model performed best in cross-validation, achieving an accuracy of 0.896, followed by XGBoost and CatBoost models with 0.891 and 0.886, respectively, while the logistic regression model showed relatively poor performance at 0.688. This suggests that the random forest, XGBoost, and CatBoost models may be more appropriate for constructing the online education adaptability prediction model. The random forest model demonstrates excellent performance in handling high-dimensional features and large-scale data, while XGBoost and CatBoost models excel in capturing complex relationships in classification problems, making them effective for prediction. Therefore, when selecting the final model, we will focus on the performance and applicability of these three models and further optimize and fine-tune them to ensure robustness and generalizability.

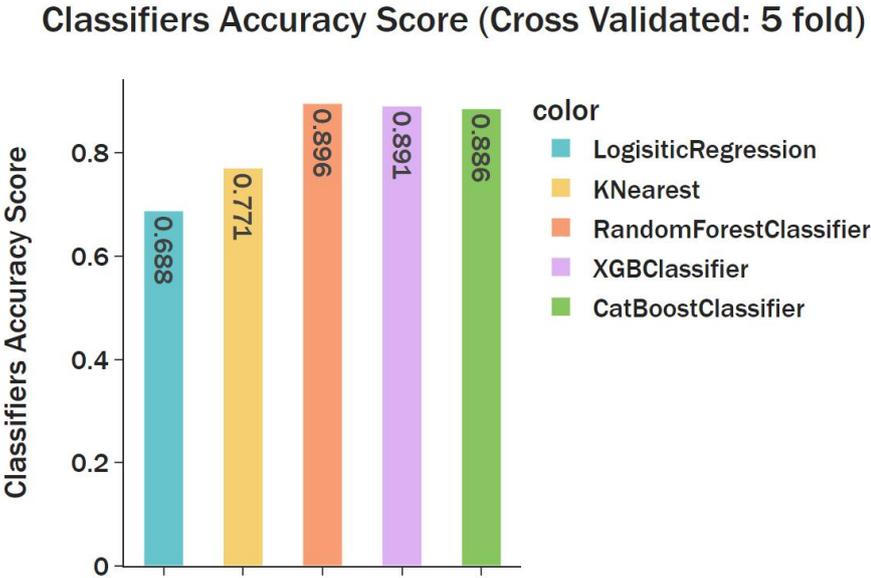

**Figure 2.** Classifiers' Accuracy Scores

## 5.Discussion and Analysis of Results

*5.1 Analysis of the Impact of Features on Adaptability*

Figure 3 presents the ranking of feature importance based on the random forest model, allowing for an analysis of the influence of various features on adaptability. The results reveal that "Class Duration," "Financial Condition," and "Age" have the most significant impact on adaptability. This suggests that the length of courses, the financial status of families, and the age of students are likely key determinants of students' adaptability in online education. Additionally, features such as gender, type of school, and type of network also have some influence on adaptability, albeit to a lesser extent.

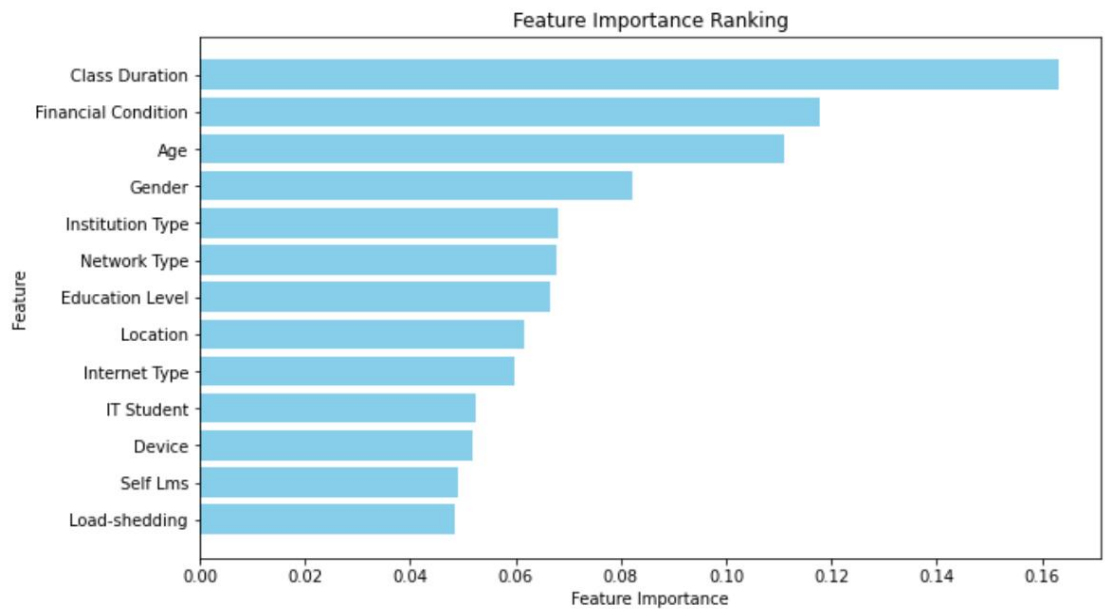

**Figure 3.** Ranking of Feature Importance

## 6. Conclusion and Prospects

This study aimed to gain an in-depth understanding of students' adaptability in online education and the factors influencing it through data analysis and machine learning model predictions. By analyzing and modeling the features in the dataset, we have reached several important conclusions:

The duration of courses, the financial status of families, and age are critical factors determining students' adaptability in online training. Longer course durations may require students to possess greater self-management skills and endurance, while better financial conditions provide students with superior learning conditions and resource support, conducive to enhancing their online learning performance and adaptability. Furthermore, age also affects students' adaptability, with older students likely having richer learning experiences and self-management skills, making them more adept at adapting to online learning environments.

Comparing the performance of different machine learning models reveals that the random forest, XGBoost, and CatBoost models perform well in predicting students.